\documentclass{article}

\usepackage{microtype}
\usepackage{graphicx}
\usepackage{subfigure}
\usepackage{booktabs} 
\usepackage{overpic}
\usepackage{hyperref}

\usepackage[accepted]{icml2023}

\usepackage{amsmath,amsfonts,bm}

\def\eqref#1{equation~\ref{#1}}

\def\1{\bm{1}}

\DeclareMathAlphabet{\mathsfit}{\encodingdefault}{\sfdefault}{m}{sl}
\SetMathAlphabet{\mathsfit}{bold}{\encodingdefault}{\sfdefault}{bx}{n}

\usepackage{amssymb}
\usepackage{amsthm}
\usepackage[utf8]{inputenc} 
\usepackage[T1]{fontenc}    
\usepackage{hyperref}      
\usepackage{url}           
\usepackage{booktabs}       
\usepackage{amsmath}       
\usepackage{nicefrac}     
\usepackage{microtype}
\usepackage{multirow}      
\usepackage{graphicx}
\usepackage{graphics}
\usepackage{tabularx}
\usepackage{makecell}
\usepackage{caption}
\usepackage{wrapfig}
\usepackage{enumitem}
\usepackage{amsfonts}      
\usepackage{nicefrac}
\usepackage{enumitem}
\usepackage{algorithmic}
\usepackage{algorithm}
\usepackage{verbatim}
\usepackage{listings}
\usepackage{wrapfig}
\usepackage{xcolor}
\usepackage{mathtools}
\definecolor{codeblue}{rgb}{0.25,0.5,0.25}
\definecolor{codekw}{rgb}{0.85, 0.18, 0.50}
\usepackage{tikz}
\usepackage{comment}

\usepackage{color, colortbl}
\definecolor{Gray}{gray}{0.93}
\definecolor{orange}{rgb}{0.9,0.5,0}

\usepackage{hyperref}
\hypersetup{colorlinks, citecolor={teal}, urlcolor=black, urlbordercolor=black}

\usepackage[margin=4pt,font=small,labelfont=bf,labelsep=endash,tableposition=bottom]{caption}
\newcolumntype{H}{>{\setbox0=\hbox\bgroup}c<{\egroup}@{}}

\usepackage[capitalize,noabbrev]{cleveref}

\theoremstyle{plain}

\theoremstyle{definition}

\theoremstyle{remark}

\usepackage[textsize=tiny]{todonotes}

\icmltitlerunning{Global Context Vision Transformers}

\begin{document}

\twocolumn[
\icmltitle{Global Context Vision Transformers}



\icmlsetsymbol{equal}{*}

\begin{icmlauthorlist}
\icmlauthor{Ali Hatamizadeh}{comp}
\icmlauthor{Hongxu Yin}{comp}
\icmlauthor{Greg Heinrich}{comp}
\icmlauthor{Jan Kautz}{comp}
\icmlauthor{Pavlo Molchanov}{comp}
\end{icmlauthorlist}
\icmlaffiliation{comp}{NVIDIA}

\icmlcorrespondingauthor{Ali Hatamizadeh}{ahatamizadeh@nvidia.com}

\icmlkeywords{Machine Learning, ICML}

\vskip 0.3in
]



\printAffiliationsAndNotice{} 

\begin{abstract}
We propose global context vision transformer (GC ViT), a novel architecture that enhances parameter and compute utilization for computer vision. Our method leverages global context self-attention modules, joint with standard local self-attention, to effectively and efficiently model both long and short-range spatial interactions, without the need for expensive operations such as computing attention masks or shifting local windows. In addition, we address the lack of the inductive bias in ViTs, and propose to leverage a modified fused inverted residual blocks in our architecture. Our proposed GC ViT achieves state-of-the-art results across image classification, object detection and semantic segmentation tasks. On ImageNet-1K dataset for classification, the variants of GC ViT with $51$M, $90$M and $201$M parameters achieve $\textbf{84.3\%}$, $\textbf{85.0\%}$ and $\textbf{85.7\%}$ Top-1 accuracy, respectively, at $224 \times 224$ image resolution and without any pre-training, hence surpassing comparably-sized prior art such as CNN-based ConvNeXt and ViT-based MaxViT and Swin Transformer by a large margin. Pre-trained GC ViT backbones in downstream tasks of object detection, instance segmentation, and semantic segmentation using MS COCO and ADE20K datasets outperform prior work consistently. Specifically, GC ViT with a 4-scale DINO detection head achieves a box AP of $\textbf{58.3\%}$ on MS COCO dataset. Code is available at
\href{https://github.com/NVlabs/GCViT}{\color{magenta}{https://github.com/NVlabs/GCViT}}.
\end{abstract}

\section{Introduction}

\begin{figure}[!ht] 
\small
\centering
    \begin{minipage}[c]{\linewidth}
    \tiny
    \begin{overpic}[width=\textwidth]
    {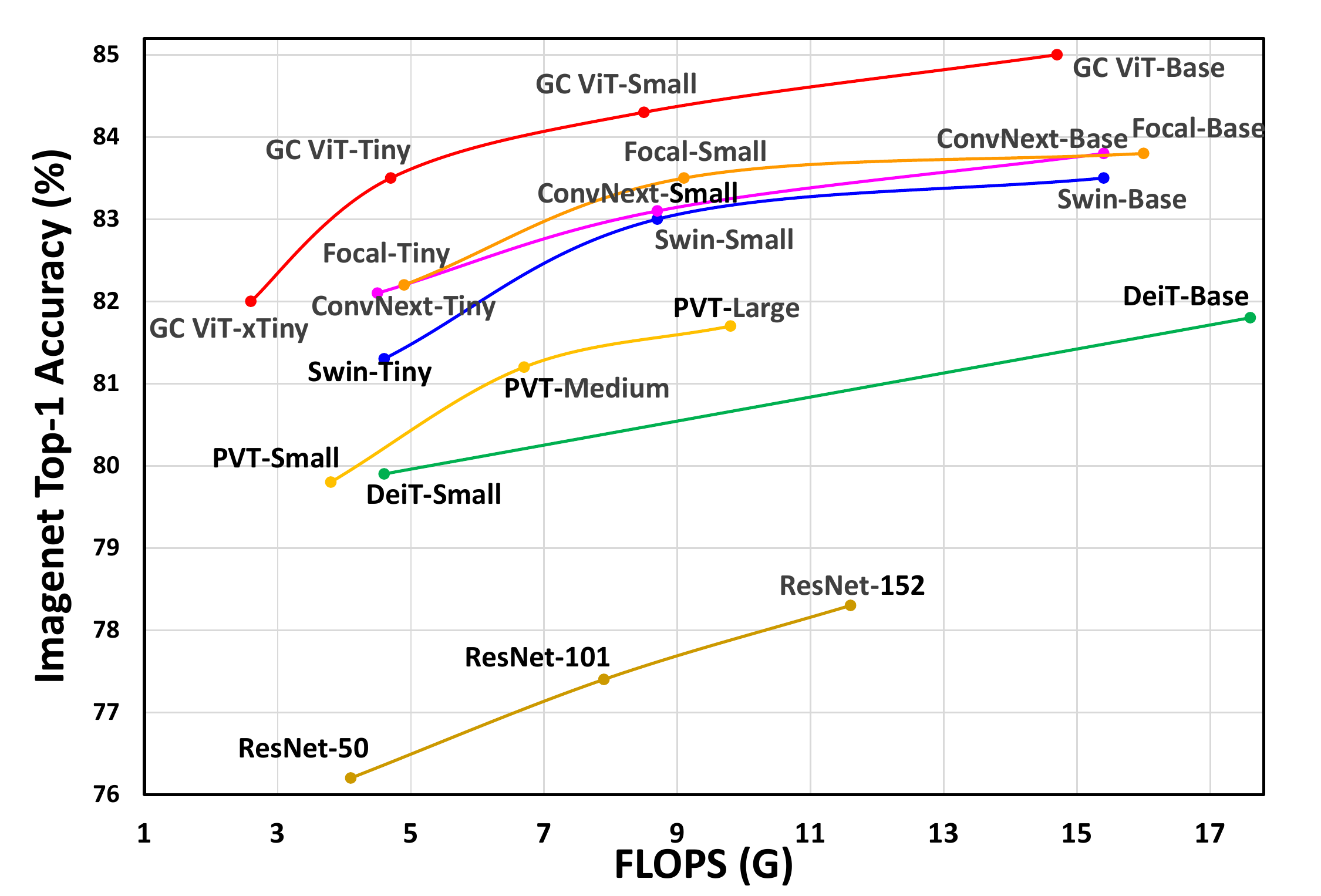}
    \end{overpic}
    \end{minipage}\hfill
    \begin{minipage}{0.328\linewidth}
        \centering
        \includegraphics[width=\textwidth]{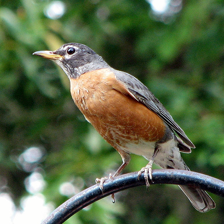}\\
        Input Image
    \end{minipage}\hfill
    \begin{minipage}{0.328\linewidth}
        \centering
        \includegraphics[width=\textwidth]{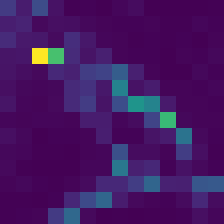}\\
        Global Attention
    \end{minipage}\hfill
    \begin{minipage}{0.328\linewidth}
        \centering
        \includegraphics[width=\textwidth]{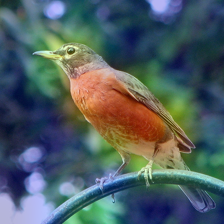}\\
        GradCAM 
    \end{minipage}
\caption{GC ViT achieves a new Pareto-front with respect to ImageNet Top-1 vs number of parameters trade-off. For fair comparison, models that are trained and evaluated with input image size of $224\times 224$ on ImageNet-1K dataset and without pre-training are considered. GC ViT is capable of capturing both short and long-range information using its global attention mechanism. We visualize corresponding attention and GradCAM maps from GC ViT to demonstrate the effectiveness of the proposed global attention mechanism.  
}
\label{fig:benchmark_fig}
\end{figure}

During the recent years, Transformers~\citep{vaswani2017attention} have achieved State-Of-The-Art (SOTA) performance in Natural Language Processing (NLP) benchmarks and became the de facto model for various tasks. A key element in the success of Transformers is the self-attention mechanism which allows for capturing contextual representations via attending to both distant and nearby tokens~\citep{yin2021adavit}. Following this trend, Vision Transformer (ViT)~\citep{dosovitskiy2020image} proposed to utilize image patches as tokens in a monolithic architecture with minor differences comparing to encoder of the original Transformer. Despite the historic dominance of Convolutional Neural Network (CNN) in computer vision, ViT-based models have achieved SOTA or competitive performance in various computer vision tasks. 

In essence, the self-attention mechanism in ViT allows for learning more uniform short and long-range information~\citep{raghu2021vision} in comparison to CNN. However, the monolithic architecture of ViT and quadratic computational complexity of self-attention baffle their swift application to high resolution images~\citep{yang2021nvit} in which capturing multi-scale long-range information is crucial for accurate representation modeling. 

\begin{figure*}[t!]
\centering
    \includegraphics[width=1.0\textwidth]{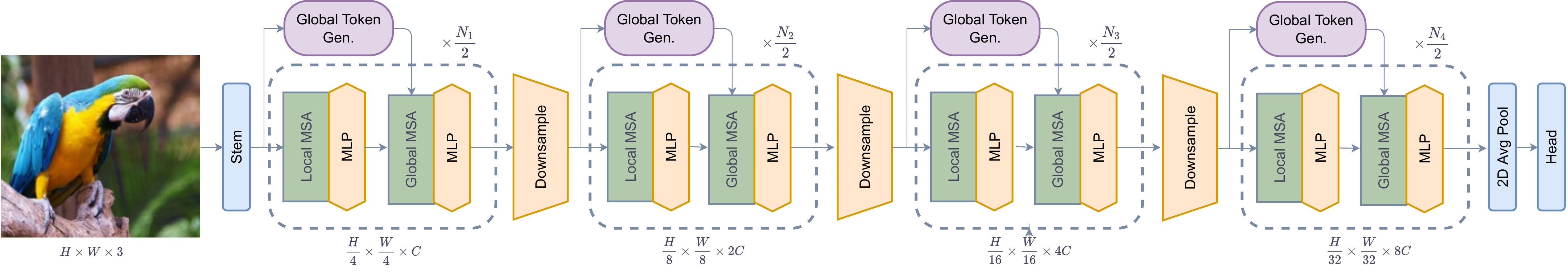}
  \caption{Architecture of the proposed GC ViT. At each stage, a query generator extracts global query tokens which captures long-range information by interacting with local key and value representations. We use alternating blocks of local and global context self attention layers. Best viewed in color.}
  \label{fig:model_architecture}
\end{figure*}

\begin{figure*}[t!]
\centering
    \includegraphics[width=0.9\textwidth,trim={35pt 10pt 15pt 25pt},clip]{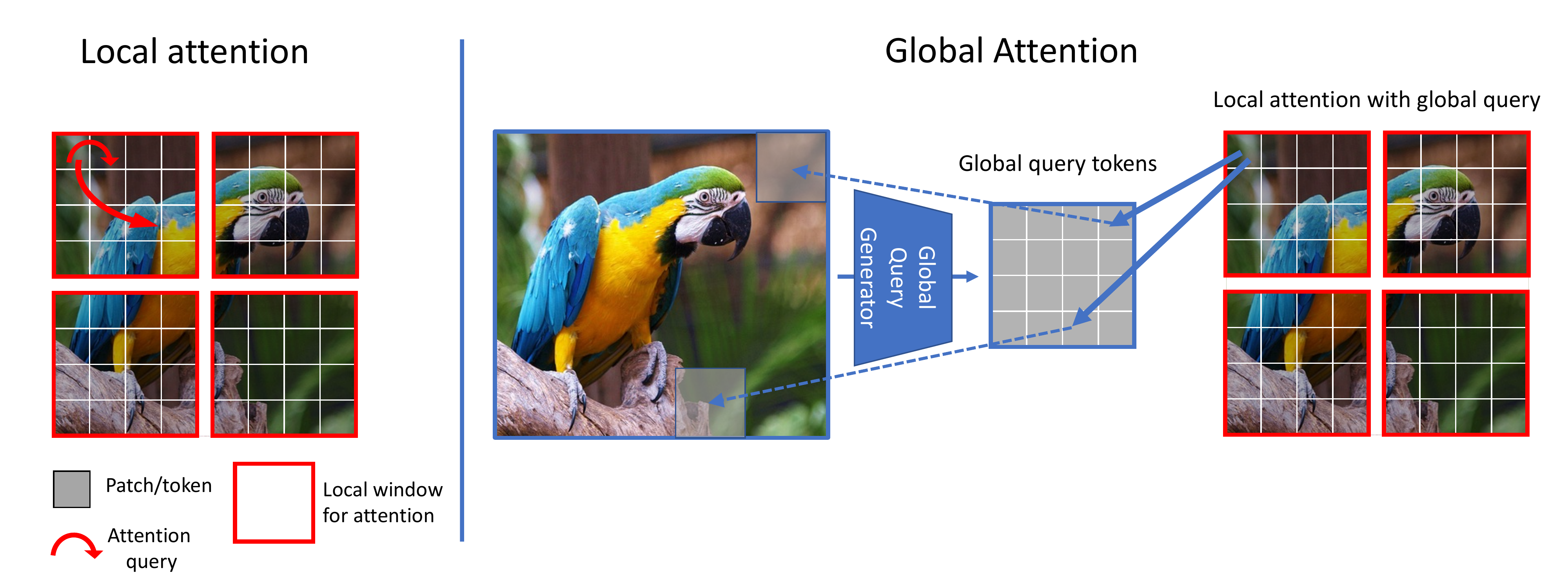}
  \caption{Attention formulation. Local attention is computed on feature patches within local window only (left). On the other hand, the global features are extracted from the entire input features and then repeated to form global query tokens. The global query is interacted with local key and value tokens, hence allowing to capture long-range information via cross-region interaction. Best viewed in color.   
  }
  \label{fig:local_global_att}
\end{figure*}

Several efforts~\citep{liu2021swin,dong2022cswin,chu2021twins,tu2022maxvit}, most notably Swin Transformer~\citep{liu2021swin}, have attempted to address the balance between short- and long-range spatial dependencies by proposing multi-resolution architectures in which the self-attention is computed in local windows. In this paradigm, cross-window connections such as window shifting are used for modeling the interactions across different regions. Despite the progress, the limited receptive field of local windows challenges the capability of self-attention to capture long-range information, and window-connection schemes such as shifting only cover a small neighborhood in the vicinity of each window. Subsequent efforts such as Focal Transformer~\citep{yang2021focal} attempted to address this issue by designing highly sophisticated self-attention modules with increased model complexity. 

In this work, we introduce the Global Context (GC) ViT network to address these limitations. Specifically, we propose a hierarchical ViT architecture consisting of local and global self-attention modules. At each stage, we compute global query tokens, using a novel fused inverted residual blocks, which we refer to as modified Fused-MBConv blocks, that encompass global contextual information from different image regions. While the local self-attention modules are responsible for modeling short-range information, the global query tokens are shared across all global self-attention modules to interact with local key and value representations. 

The design of our proposed framework for global query generator and self-attention is intuitive and simple and can be efficiently implemented using major deep learning framework. Hence, it eliminates sophisticated and computationally expensive operations and ensures the effectiveness of self-attention when applied to high-resolution images. In addition, we propose a novel downsampling block with a parameter-efficient fused-MBConv layer to address the lack of inductive bias in ViTs and enhancing the modeling of inter-channel dependencies.  

We have extensively validated the effectiveness of the proposed GC ViT using three publicly available datasets for various computer vision tasks. For image classification using ImageNet-1K dataset, GC ViT with $51$M, $90$M, $201$M parameters achieve new SOTA benchmarks of $\textbf{84.3\%}$, $\textbf{85.0\%}$, $\textbf{85.7\%}$ Top-1 accuracy and without using extra data or pre-training. 

Hence, GC ViT consistently outperforms both ConvNeXt \citep{liu2022convnet}, MaxViT~\citep{tu2022maxvit} and Swin Transformer \citep{liu2021swin} models, sometimes by a significant margin (see Fig.~\ref{fig:benchmark_fig}). 

Using an ImageNet-1K pre-trained GC ViT base backbone with a Cascade Mask RCNN~\citep{he2017mask} head, our model achieves a box mAP of \textbf{52.9} for object detection and a mask mAP of \textbf{45.8} for instance segmentation on the MS COCO dataset and by using single-scale inference. We also used an ImageNet-21K GC ViT model as backbone with a 4-scale DINO detection head and achieved a box AP of $\textbf{58.3\%}$. 

In addition, using an UPerNet \citep{xiao2018unified} head, our model achieves a mIoU of \textbf{49.2} on ADE20K for semantic segmentation by only using a single-scale inference scheme. Other variants of GC ViT with different learning capacities also demonstrate SOTA results when compared to similarly-sized models on both MS COCO and ADE20K datasets. Hence, GC ViT demonstrates great scalability for high-resolution images on various downstream tasks, validating the effectiveness of the proposed framework in capturing both short and long-range information.

The main contributions of our work are summarized as follows:

\begin{itemize}[leftmargin=*,nosep]
\item We introduce a compute and parameter-optimized hierarchical ViT with reparametrization of the design space (\textit{e.g.}, embedding dimension, number of heads, MLP ratio).
\item We design an efficient CNN-like token generator that encodes spatial features at different resolutions for global query representations.

\item We propose global query tokens that can effectively capture contextual information in an efficient manner and model both local and global interactions.
\item We introduce a parameter-efficient downsampling module with modified Fused MB-Conv blocks that not only integrates inductive bias but also enables the modeling of inter-channel dependencies.

\item We demonstrate new SOTA benchmarks for : (1) ImageNet classification with Pareto fronts on ImageNet-1K for number of parameters and FLOPs (2) downstream tasks such as detection, instance segmentation and semantic segmentation on MS COCO and ADE20K, respectively.
\end{itemize}
\section{GC ViT architecture}

\label{sec:arch}
\textbf{Architecture.} Fig.~\ref{fig:model_architecture} depicts the architecture of GC ViT. We propose a hierarchical framework to obtain feature representations at several resolutions (called stages) by decreasing the spatial dimensions while expanding the embedding dimension, both by factors of $2$. 

At first, given an input image with resolution of $\ \mathbf{x}\in \mathbb{R}^{H\times W\times 3}$, we obtain overlapping patches by applying a $3 \times 3$ convolutional layer with a stride of $2$ and appropriate padding. Then patches are projected into a $C$-dimensional embedding space with another $ 3 \times 3$ convolutional layer with stride $2$. 

Every GC ViT stage is composed of alternating local and global self-attention modules to extract spatial features. Both operate in local windows like Swin Transformer~\citep{liu2021swin}, however, the global self-attention has access to global features extracted by the global query generator. The query generator is a CNN-like module that extracts features from the entire image only once at every stage. After each stage, the spatial resolution is decreased by $2$ while the number of channels is increased by $2$ via a downsampling block. Resulting features are passed through average pooling and linear layers to create an embedding for a downstream task.

The GC ViT architecture benefits from novel blocks such as \textit{a downsampling operator}, \textit{a global query generator} and \textit{a global self-attention module} described in the next sections.

\textbf{Downsampler.} We leverage an idea of spatial feature contraction from CNN models that imposes locality bias and cross channel interaction while reducing dimensions.  We utilize a modified Fused-MBConv block, followed by a max pooling layer with a kernel size of $3$ and stride of $2$ as a downsampling operator. The Fused-MBConv block in our work is similar to the one in EfficientNetV2~\citep{tan2021efficientnetv2} with modifications as in
\begin{align}
\begin{split}
\mathbf{\hat{x}} & = \text{DW-Conv}_{3\times3}(\mathbf{x}), \\
\mathbf{\hat{x}} & = \text{GELU}(\mathbf{\hat{x}}), \\
\mathbf{\hat{x}} & = \text{SE}(\mathbf{\hat{x}}), \\
\mathbf{x} & = \text{Conv}_{1\times1}(\mathbf{\hat{x}}) + \mathbf{x},
\label{eq:fused_convmb}
\end{split}
\end{align}
where $\text{SE}$, $\text{GELU}$ and $\text{DW-Conv}_{3\times3}$ denote Squeeze and Excitation block~\citep{hu2018squeeze}, Gaussian Error Linear Unit~\citep{hendrycks2016gaussian} and $3\times3$ depth-wise convolution, respectively. In our proposed architecture, the Fused-MBConv blocks provide desirable properties such as inductive bias and modeling of inter-channel dependencies.  It is ablated in Table~\ref{tab:abl-sup-downsampler}.

\begin{figure*}[t!]
\centering%
    \includegraphics[width=0.9\textwidth,trim={20pt 12pt 0pt 15pt}, clip]{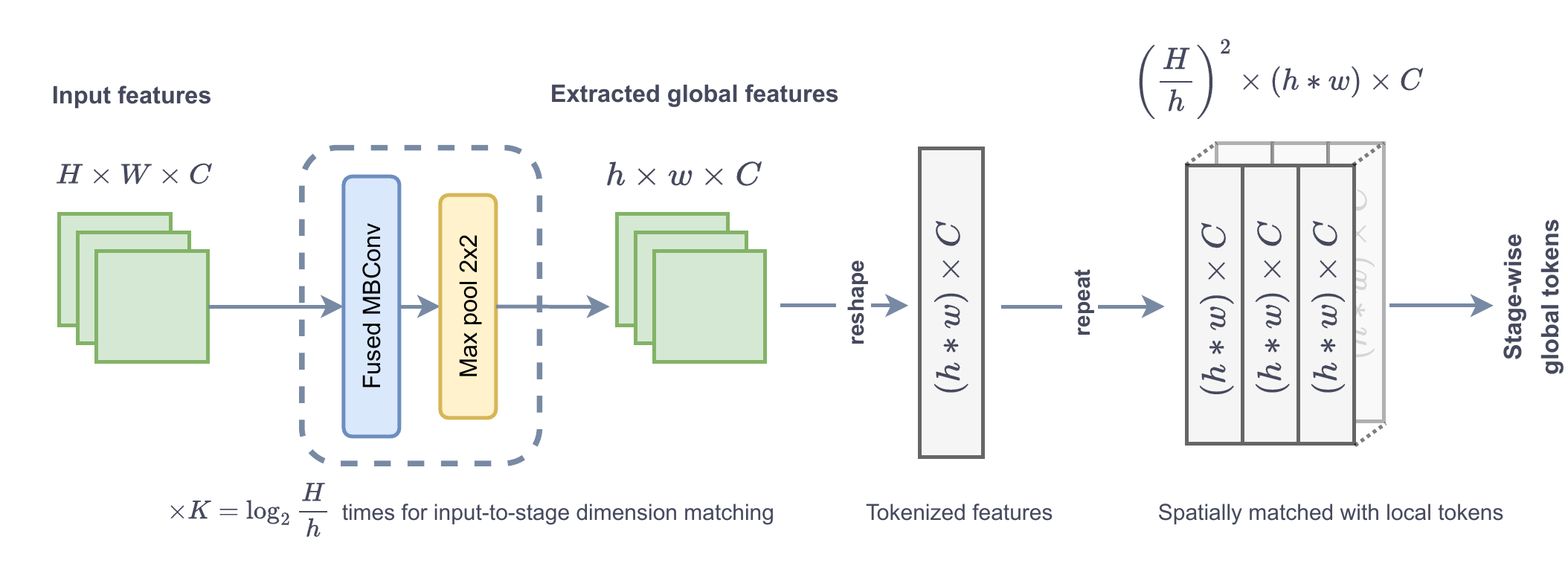}
  \caption{
  Global query generator schematic diagram. It is designed to (i) transform an input feature map to the current stage of dimension $H, W, C$ denoting height, width, and channel respectively, (ii) extract features via repeating the modified Fused MBConv block, joint with down-sampling, $\text{log}_{2}\frac{H}{h}$ times for dimension matching to local window size $h$ (iii) output is reshaped and repeated to $(\frac{H}{h})^2$ number of local tokens that can attend to global contextual information. $\star$ denotes merged dimensions during reshaping.
  }
  \label{fig:global_query}
\end{figure*}

\subsection{Global Self-Attention}
\label{sec:global_sa}
Fig.~\ref{fig:local_global_att} demonstrates the main idea behind our contribution. Local self-attention can only query patches within a local window, whereas the global attention can query different image regions while still operating within the window. At each stage, the global query component is pre-computed.
The global self-attention utilizes the extracted global query tokens
and shared across all blocks, to interact with the local key and value representations. In addition, GC ViT employs alternating local and global self-attention blocks to effectively capture both local and global spatial information. Fig.~\ref{fig:attention_blocks} illustrates the difference between local and global self-attention. The global attention query $\mathbf{q_g}$ has a size of $B \times C \times h \times  w$, wherein $B$, $C$, $h$ and $w$ denote batch size, embedding dimension, local window height and width, respectively. Moreover, $\mathbf{q_g}$ is repeated along the batch dimension to compensate for the overall number of windows and aggregated batch size $B^* = B \times N^*$ where $N^*$ is the number of local windows. $\mathbf{q_g}$ is further reshaped into multiple heads. The value and key are computed within each local window using a linear layer.

\begin{minipage}{0.48\textwidth}
\floatname{algorithm}{Algorithm.}
\begin{algorithm}[H]
\caption{Global Attention Pseudocode}
\label{alg:global_code}
\begin{lstlisting}[language=python]
# Input/output shape: (B*, N, C);
# B*: Aggregated Batch Size; H: Height;
# W: Width; C: dim; q_g: Global Token;
# F: Num Attention Head; N: H x W.
def init():
    f = nn.Linear(C, 2*C)
    softmax = nn.Softmax(dim=-1)

def forward(x, q_g):
    B*, N, C = x.shape
    B, C, h, w = q_g.shape
    kv = f(x).reshape(B*, N, 2, F, C // F)
    kv = kv.permute(2, 0, 3, 1, 4)
    k, v = split(kv, (1, 1), 0)
    q_g = q_g.repeat(1, B* // B, 1, 1)
    q_g = q_g.reshape(B*, F, N, C // F)
    qk = matmul(q_g,k.transpose(-2, -1))
    attn = softmax(qk)
    return matmul(attn, v).reshape(B*, N, C)
\end{lstlisting}
\end{algorithm}
\end{minipage}


Since the partitioned windows only contain local information, interaction with rich contextual information embedded in the global query tokens provides an effective way of enlarging the receptive field and attending to various regions in the input feature maps. The self-attention module is computed as in    
\begin{equation}
\small
    \text{Attention}(\mathbf{q_{g}}, \mathbf{k}, \mathbf{v}) = \text{Softmax}(\frac{\mathbf{q_{g}} \mathbf{k}}{\sqrt{\mathbf{d}}} + \mathbf{b})\mathbf{v},
    \label{eq:kv}
\end{equation}



where $\mathbf{d}$ is scaling factor and $\mathbf{b}$ is a learnable relative position bias term. Assuming position change between $[-p+1, p-1]$ along horizontal and vertical axes,  $\mathbf{b}$ is sampled from the grid $\hat{\mathbf{b}} \in \mathbb{R}^{(2p-1) \times (2p-1)}$. As shown in Sec.~\ref{sec:abl}, relative position bias improves the performance, especially for dense prediction downstream tasks. In Algorithm~\ref{alg:global_code}, we present a PyTorch-like pseudocode for computing global self-attention in GC ViT.

\subsection{Complexity Analysis}
Given an input feature map of $x \in \mathcal{R}^{H \times W \times C}$ at each stage with a window size of $h \times w$, the computational complexity of GC ViT is as follows
\begin{equation}
\small
    \mathcal{O}(\text{GC ViT}) = 2HW(2C^{2} + hwC),
    \label{eq:fsa}
\end{equation}
The efficient design of global query token generator and other components allows to maintain a similar computational complexity in comparison to Swin Transformer~\cite{liu2021swin} while being able to capture long-range information and achieve better higher accuracy for classification and downstream tasks such as detection and segmentation.
\section{Experiments}
\label{sec:exp}

\begin{table}[!t]
	\caption{Image classification benchmarks on \textbf{ImageNet-1K} dataset~\citep{deng2009imagenet}. Models that are trained on ImageNet-1K dataset and without any pre-training or usage of extra data are considered.}
	\label{tab:classfication_main}
	\centering
    \resizebox{1\linewidth}{!}{
	\begin{tabular}[t]{ l |c|c|c|Hl}
		\toprule
		Model & Param (M) & FLOPs (G) & Image Size & Step (s)& Top-1 (\%)  \\
          \midrule
		\multicolumn{6}{c}{ConvNet} \\
   \midrule
   ResNet50~\citep{he2016deep}   &25 &4.1 &224$^{2}$ &-&76.1 \\
   ResNet-101~\citep{he2016deep}  &44 & 7.9 &224$^{2}$&-&77.4\\
   ResNet-152~\citep{he2016deep} &60 & 11.6 &224$^{2}$ &-&78.3 \\
EfficientNetV2-B2~\cite{tan2021efficientnetv2}  & 10 & 1.6 & 260$^{2}$&- & 80.2\\
EfficientNetV2-B3~\cite{tan2021efficientnetv2}  & 14 & 2.9 & 300$^{2}$ &- & 82.0\\
EfficientNetV2-S~\cite{tan2021efficientnetv2}  & 21 & 8.0 & 384$^{2}$ &- & 83.9\\
RegNetY-040~\cite{radosavovic2020designing}  & 20 & 6.6 & 288$^{2}$&-& 83.0\\
RegNetY-064~\cite{radosavovic2020designing}  & 30 & 10.5 & 288$^{2}$&- & 83.7\\
   ConvNeXt-T~\citep{liu2022convnet}  & 29& 4.5 & 224$^{2}$ & 0.90 & 82.1 \\
   ConvNeXt-S~\citep{liu2022convnet}  & 50& 8.7 & 224$^{2}$ & 0.90 & 83.1 \\
   ConvNeXt-B~\citep{liu2022convnet}  & 89& 15.4 & 224$^{2}$ & 0.90 & 83.8\\
   ConvNeXt-L~\citep{liu2022convnet}  & 198& 34.4 & 224$^{2}$ & 0.90 & 84.3\\
          \midrule
		\multicolumn{6}{c}{Transformer} \\
		\midrule
  ViT-B~\citep{dosovitskiy2020image} & 86 & 17.6 & 224$^{2}$& & 77.9 \\
  DeiT-S/16~\citep{touvron2021training}  & 22 & 4.6 & 224$^{2}$ && 79.9 \\
		DeiT-B~\citep{touvron2021training} & 86 & 17.6 & 224$^{2}$& & 81.8 \\
	
Swin-T \citep{liu2021swin} & 29 & 4.5& 224$^{2}$ & 0.48 &81.3 \\
		Swin-S~\citep{liu2021swin} & 50 & 8.7 & 224$^{2}$& 0.73 &83.0\\
  Swin-B \citep{liu2021swin} & 88 & 15.4& 224$^{2}$& 0.89 &83.3  \\
  Twins-S~\cite{chu2021twins}  & 24 & 2.8 & 224$^{2}$&- & 81.7\\
Twins-B~\cite{chu2021twins}  & 56 & 8.3 & 224$^{2}$&- & 83.1\\
Twins-L~\cite{chu2021twins}  & 99 & 14.8 & 224$^{2}$&- & 83.7\\
   Focal-T~\citep{yang2021focal}  & 29& 4.9 & 224$^{2}$ & 0.90 & 82.2
		\\
		Focal-S~\citep{yang2021focal}  & 51& 9.1 & 224$^{2}$ & 0.90 & 83.5 \\
  Focal-B~\citep{yang2021focal}  & 90& 16.0 & 224$^{2}$ & 0.90 & 83.8
		\\ 
 PoolFormer-S36~\cite{yu2022metaformer}  & 31 & 5.0 & 224$^{2}$&- & 81.4\\
PoolFormer-M36~\cite{yu2022metaformer}  & 56 & 8.8 & 224$^{2}$&- & 82.1\\ 
PoolFormer-M58~\cite{yu2022metaformer}  & 73 & 11.6 & 224$^{2}$&- & 82.4\\
 SwinV2-T~\cite{liu2022swin}  & 28 & 4.4 & 256$^{2}$&- & 81.8\\
SwinV2-S~\cite{liu2022swin}  & 49 & 8.5 & 256$^{2}$&- & 83.8\\
SwinV2-B~\cite{liu2022swin}  & 88 & 15.1 & 256$^{2}$&- & 84.6\\

 \midrule
		\multicolumn{6}{c}{Hybrid} \\
		\midrule
  CrossViT-S~\cite{chen2021crossvit}  & 27 & 5.1 & 224$^{2}$&- & 81.0\\
CrossViT-B~\cite{chen2021crossvit}  & 105 & 20.1 & 224$^{2}$&- & 82.2\\
CoAtNet-0~\citep{dai2021coatnet}  & 25& 4.2 & 224$^{2}$ & 0.90 & 81.6\\
  CoAtNet-1~\citep{dai2021coatnet}  & 42& 8.4 & 224$^{2}$ & 0.90 & 83.3\\
  CoAtNet-2~\citep{dai2021coatnet}  & 42& 8.4 & 224$^{2}$ & 0.90 & 83.3\\
		CoAtNet-3~\citep{dai2021coatnet}  & 168& 34.7 & 224$^{2}$ & 0.90 & 84.5\\
  PVT-v2-B2~\citep{wang2022pvt}   &25 &4.0 &224$^{2}$&-&82.0 \\
 		PVT-v2-B3~\citep{wang2022pvt}   &45 &6.9 &224$^{2}$&-&83.2 \\
		PVT-v2-B5~\citep{wang2022pvt}   &82 &11.8 &224$^{2}$&-&83.8 \\
    CSwin-T~\citep{dong2022cswin}  & 23& 4.3 & 224$^{2}$ & 0.90 & 82.7
 \\
		CSwin-S~\citep{dong2022cswin}  & 35& 6.9 & 224$^{2}$ & 0.90 & 83.6\\
  CSwin-B~\citep{dong2022cswin}  & 78& 15.0 & 224$^{2}$ & 0.90 & 84.2
		\\
    MaxViT-T~\citep{tu2022maxvit}  & 31& 5.6 & 224$^{2}$ & 0.90 & 83.6
    		\\
    MaxViT-S~\citep{tu2022maxvit}  & 69& 11.7 & 224$^{2}$ & 0.90 & 84.4
    		\\
    MaxViT-B~\citep{tu2022maxvit}  & 120& 74.2 & 224$^{2}$ & 0.90 & 84.9
    		\\
    MaxViT-L~\citep{tu2022maxvit}  & 212& 43.9 & 224$^{2}$ & 0.90 & 85.1
 \\
            \midrule
		\multicolumn{6}{c}{\textbf{GC ViT}} \\
   \midrule
  \rowcolor{Gray}
		\textbf{GC ViT-XXT}& 12& 2.1 & 224$^{2}$ & 0.90 & \textbf{79.9}  \\
  \rowcolor{Gray}
		\textbf{GC ViT-XT}& 20& 2.6 & 224$^{2}$ & 0.90 & \textbf{82.0}  \\ 
		\rowcolor{Gray}
		\textbf{GC ViT-T}& 28& 4.7 & 224$^{2}$ & 0.90 & \textbf{83.5}\\ 
  		\rowcolor{Gray}
		\textbf{GC ViT-T2}& 34& 5.5 & 224$^{2}$ & 0.90 & \textbf{83.7}\\\rowcolor{Gray}
		\textbf{GC ViT-S}& 51& 8.5 & 224$^{2}$ & 0.90 & \textbf{84.3} \\
  		\rowcolor{Gray}
		\textbf{GC ViT-S2}& 68& 10.7 & 224$^{2}$ & 0.90 & \textbf{84.8}\\
   \rowcolor{Gray}
		\textbf{GC ViT-B}& 90& 14.8 & 224$^{2}$ & 0.90 & \textbf{85.0} \\
		\rowcolor{Gray}
		\textbf{GC ViT-L}& 201& 32.6 & 224$^{2}$ & 0.90 & \textbf{85.7} \\
		\bottomrule

	\end{tabular}
	}
\label{tab:imgnet}
\end{table}

For image classification, we trained and tested our model on ImageNet-1K dataset \citep{deng2009imagenet}. To allow for a fair comparison, all GC ViT variants are trained by following training configurations of previous efforts \citep{liu2021swin,yang2021focal,chu2021twins}. Specifically, all models are trained with the AdamW \citep{kingma2014adam} optimizer for $300$ epochs with an initial learning rate of $0.001$, weight decay of $0.05$, cosine decay scheduler and 20 warm-up and cool-down epochs, respectively.

For object detection and instance segmentation, we trained our model on MS COCO \citep{lin2014microsoft} with DINO~\citep{he2017mask} and a Mask-RCNN~\citep{he2017mask} heads, using $\times3$ LR schedule with an initial learning rate of $0.0001$, a batch size of $16$ and weight decay of $0.05$. Following~\cite{liu2022convnet}, we compared against Tiny, Small and Base model variants using Cascade Mask-RCNN but only compared against Tiny variant using Mask-RCNN. For semantic segmentation, we used the ADE20K dataset \citep{zhou2017scene} with a UPerNet~\citep{xiao2018unified} segmentation head. Following previous efforts, we used a random crop size of $512\times512$ for the input images.

\setlength{\tabcolsep}{8pt}
\begin{table*}[!t]
    \centering
    \caption{Object detection and instance segmentation benchmarks using Mask R-CNN and Cascade Mask R-CNN on \textbf{MS COCO} dataset~\citep{lin2014microsoft}. All models employ $3\times$ schedule.}
    \label{tab:cascademaskrcnn}
    \resizebox{0.95\linewidth}{!}{
    \begin{tabular}{l|cc|cccccc}
        \toprule
        Backbone & Param (M) & FLOPs (G) & $\text{AP}^{\text{box}}$ & $\text{AP}^{\text{box}}_{50}$ & $\text{AP}^{\text{box}}_{75}$ & $\text{AP}^{\text{mask}}$ & $\text{AP}^{\text{mask}}_{\text{50}}$ & $\text{AP}^{\text{mask}}_{\text{75}}$ \\ 
        \midrule
		\multicolumn{9}{c}{Mask-RCNN 3$\times$ schedule} \\
		\midrule

        Swin-T~\citep{liu2021swin} & 48 & 267 & 46.0 & 68.1 & 50.3 & 41.6 & 65.1 & 44.9 \\
        ConvNeXt-T~\citep{liu2022convnet} & 48 & 262 & 46.2 & 67.9 & 50.8 & 41.7 & 65.0 &44.9
        \\ \rowcolor{Gray}
        \textbf{GC ViT-T} & 48 & 291 & \textbf{47.9} & \textbf{70.1} & \textbf{52.8} & \textbf{43.2} & \textbf{67.0} & \textbf{46.7} \\
        \midrule
		\multicolumn{9}{c}{Cascade Mask-RCNN 3$\times$ schedule} \\
		\midrule
        DeiT-Small/16~\citep{touvron2021training} & 80 & 889 & 48.0 & 67.2 & 51.7 & 41.4 & 64.2 & 44.3 \\
		ResNet-50~\citep{he2016deep} & 82 & 739 & 46.3 & 64.3 & 50.5 & 40.1 & 61.7 & 43.4 \\
        Swin-T~\citep{liu2021swin} & 86 & 745 & 50.4 & 69.2 & 54.7 & 43.7 & 66.6 & 47.3 \\
        ConvNeXt-T~\citep{liu2022convnet} & 86 & 741 & 50.4 & 69.1 & 54.8 & 43.7 & 66.5 & 47.3
        \\ \rowcolor{Gray}
        \textbf{GC ViT-T} & 85 & 770 & \textbf{51.6} & \textbf{70.4} & \textbf{56.1} & \textbf{44.6} & \textbf{67.8} & \textbf{48.3} \\
		\midrule
		X101-32~\citep{xie2017aggregated} & 101 & 819 & 48.1 & 66.5 & 52.4 & 41.6 & 63.9 & 45.2 \\
        Swin-S~\citep{liu2021swin} & 107 & 838  & 51.9 & 70.7 & 56.3 & 45.0 & 68.2 & 48.8 \\
        ConvNeXt-S~\citep{liu2022convnet} & 108 & 827 & 51.9 & 70.8 & 56.5 & 45.0 & 68.4 & 49.1
        \\ \rowcolor{Gray}
        \textbf{GC ViT-S} & 108 & 866 & \textbf{52.4} & \textbf{71.0} & \textbf{57.1} & \textbf{45.4} & \textbf{68.5} & \textbf{49.3} \\
        \midrule
        X101-64~\citep{xie2017aggregated} & 140 & 972 & 48.3 & 66.4 & 52.3 & 41.7 & 64.0 & 45.1 \\
        Swin-B~\citep{liu2021swin} & 145 & 982  & 51.9 & 70.5 & 56.4 & 45.0 & 68.1 & 48.9
        \\
        ConvNeXt-B~\citep{liu2022convnet} & 146 & 964 & 52.7 & 71.3 & 57.2 & 45.6 & 68.9 & 49.5 \\
        \rowcolor{Gray}
        \textbf{GC ViT-B} & 146 & 1018 & \textbf{52.9} & \textbf{71.7} & \textbf{57.8} & \textbf{45.8} & \textbf{69.2} & \textbf{49.8} \\
        \bottomrule
    \end{tabular}
    }
\end{table*}

\subsection{Classification}
\label{sec:exp_imagenet_results}

We present the ImageNet-1K classification benchmarks in Table~\ref{tab:imgnet} and compare against CNN and ViT-based models across different model sizes. Our model achieves better performance when compared to other established benchmarks such as ConvNeXt~\citep{liu2022convnet}. Furthermore, as shown in Fig.~\ref{fig:benchmark_fig}, GC ViT models have better or comparable computational efficiency in terms of number FLOPsover the competing counterpart models.

\subsection{Detection and Instance Segmentation}
\label{sec:exp_mscoco_results}
In Table~\ref{tab:cascademaskrcnn}, we present object detection and instance segmentation benchmarks on MS COCO dataset. Using a Mask-RCNN head, the model with pre-trained GC ViT-T (47.9/43.2) backbone outperforms counterparts with pre-trained ConvNeXt-T \citep{liu2022convnet} (46.2/41.7) by +1.7 and +1.5 and Swin-T \citep{liu2021swin} (46.0/41.6) by +1.9 and +1.6 in terms of box AP and mask AP, respectively. Using a Cascade Mask-RCNN head, the models with pre-trained GC ViT-T (51.6/44.6) and GC ViT-S (52.4/45.4) backbones outperform ConvNeXt-T \citep{liu2022convnet} (50.4/43.7) by +1.2 and +0.9 and ConvNeXt-S \citep{liu2022convnet} (51.9/45.0) by +0.5 and +0.4 in terms of box AP and mask AP, respectively. Furthermore, the model with GC ViT-B (52.9/45.8) backbone outperforms the counterpart with ConvNeXt-B \citep{liu2022convnet} (52.7/45.6) by +0.2 and +0.2 in terms of box AP and mask AP, respectively. 

As shown in Table~\ref{tab:cascademaskrcnn}, we have also tested the performance of GC ViT-L model, pre-trained on ImageNet-21K dataset, with a 4-scale DINO~\citep{zhang2022dino} detection head and achieved a box AP of $\textbf{58.3\%}$ on MS COCO dataset. Hence our model outperforms the counterpart with Swin-L backbone.

\subsection{Semantic Segmentation}
\label{sec:exp_ade_seg_results}
We present semantic segmentation benchmarks on ADE20K dataset in Table~\ref{tab:ade_segmentation}. The models using pre-trained GC ViT-T (47.0), GC ViT-S (48.3) and GC ViT-B (49.2) backbones outperform counterpart models with pre-trained Twins-SVT-S \citep{chu2021twins} (46.2), Twins-SVT-B \citep{chu2021twins} (47.7) and Twins-SVT-L \citep{chu2021twins} (48.8) by +0.8, +0.6 and +0.4 in terms of mIoU, respectively. In addition, models with GC ViT backbones significantly outperform counterparts with Swin Transformer backbones, hence demonstrating the effectiveness of the global self-attention.

\begin{table}
\centering
\resizebox{1.0\linewidth}{!}{
\footnotesize
\setlength{\tabcolsep}{2.5pt}
  \begin{tabular}{lccc}
    \toprule
    Backbone  & Head & Scale & $\text{AP}^{\text{box}}$ \\
    \midrule
    ResNet-50~\citep{he2016deep}&  DINO~\citep{zhang2022dino}&4 & 50.9 \\
    ResNet-50~\citep{he2016deep}&  DINO~\citep{zhang2022dino}&5 & 51.2 \\
    Swin-L$^\ddag$~\citep{liu2021swin}  & DINO~\citep{zhang2022dino} & 4 & 58.0\\
    \rowcolor{Gray}
    \textbf{GC ViT-L}$^\ddag$  & DINO~\citep{zhang2022dino} & 4 & \textbf{58.3}\\
    \bottomrule

  \end{tabular} 
  }
    \caption{Object detection benchmarks using DINO~\citep{zhang2022dino} network on \textbf{MS COCO} dataset~\citep{lin2014microsoft}. $^\ddag$ denotes models that are pre-trained on ImageNet-21K dataset.}
    \label{tab:dino}
\end{table}

\begin{table}
\centering
\resizebox{1.0\linewidth}{!}{
\footnotesize
\setlength{\tabcolsep}{2.5pt}
  \begin{tabular}{lccc}
    \toprule
    Backbone  & Param (M) & FLOPs (G) & mIoU \\
    \midrule	 
    DeiT-Small/16~\citep{touvron2021training}  & 52 & 1099 & 44.0\\
    Swin-T~\citep{liu2021swin}  & 60 & 945 & 44.5\\
    ResNet-101~\citep{he2016deep}  & 86 & 1029 & 44.9\\
    Focal-T~\citep{yang2021focal}  & 62 & 998 & 45.8\\
    Twins-SVT-S~\citep{chu2021twins}  & 55 & - & 46.2\\
    \rowcolor{Gray}
    \textbf{GC ViT-T}  & 58 & 947 & \textbf{47.0}\\
    \midrule
    Swin-S~\citep{liu2021swin}  & 81 & 1038 & 47.6\\
    Twins-SVT-B~\citep{chu2021twins}  & 89 & - & 47.7\\
    Focal-S~\citep{yang2021focal}  & 85 & 1130 & 48.0\\
    \rowcolor{Gray}
    \textbf{GC ViT-S}  & 84 & 1163 & \textbf{48.3}\\
    \midrule
    Swin-B~\citep{liu2021swin}  & 121 & 1188 & 48.1\\
    Twins-SVT-L~\citep{chu2021twins}  & 133 & - & 48.8\\
    Focal-B~\citep{yang2021focal}  & 126 & 1354 & 49.0\\
    \rowcolor{Gray}
    \textbf{GC ViT-B}  & 125 & 1348 & \textbf{49.2}\\
    \bottomrule

  \end{tabular} 
  }
    \caption{Semantic segmentation benchmarks \textbf{ADE20K} validation set with UPerNet~\citep{xiao2018unified} and pre-trained ImageNet-1K backbone. All models use a crop size of $512\times512$ and use single-scale inference.}
    \label{tab:ade_segmentation}
\end{table}

\begin{figure*}[!t]
\centering

\resizebox{0.85\linewidth}{!}{
\begingroup
\renewcommand*{\arraystretch}{0.3}
\begin{tabular}{c}

  \includegraphics[width=1\linewidth]{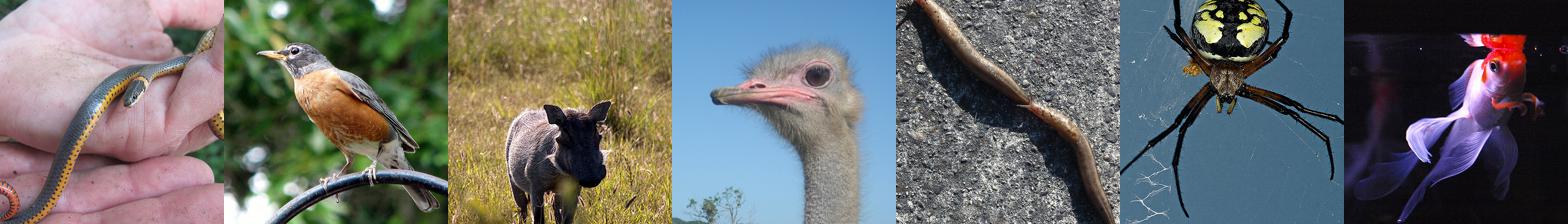} \\ 
  \small{(a) Original images from ImageNet-1K validation set.} \\[3pt]
 
\includegraphics[width=1\linewidth]{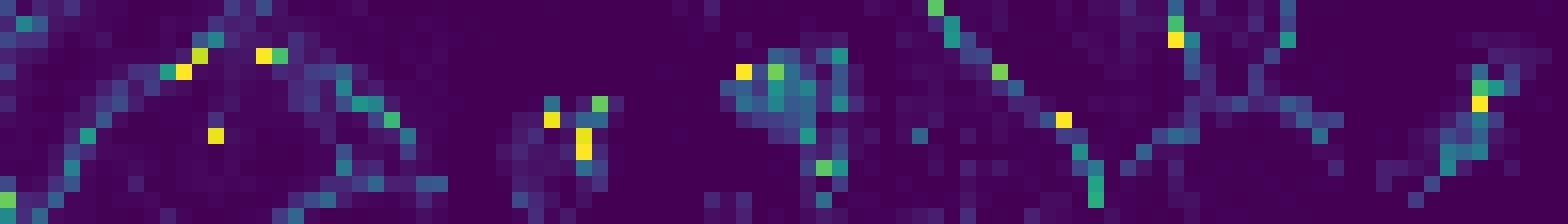} \\
 \small{(b) \textbf{Global attention} maps from GC ViT model (ours).}
 \\ 
 
 \includegraphics[width=1\linewidth]{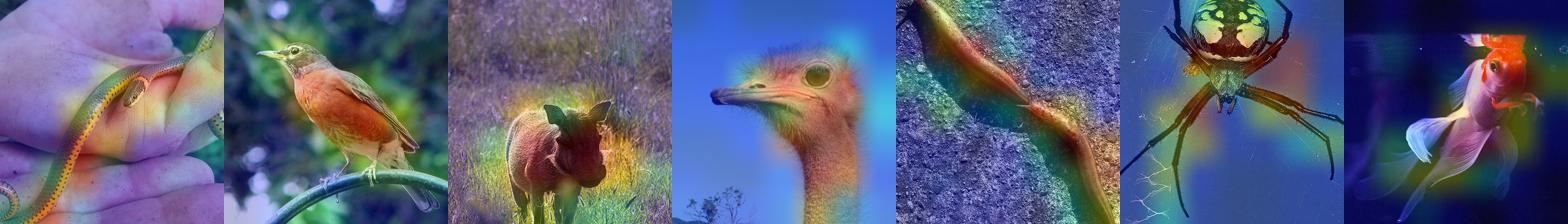} \\
 \small{(c) Corresponding \textbf{Grad-CAM} maps.}

\end{tabular}
\endgroup
}
\caption{Visualization of : (a) input images (b) global self-attention maps from GC ViT-T model (c) corresponding Grad-CAM attention maps. Both short and long-range spatial dependencies are captured effectively.}
\vspace{-2mm}
\label{fig:att_maps}
\end{figure*}

\section{Ablation}
\label{sec:abl}


\begin{table}
\resizebox{1\linewidth}{!}{
\centering
\footnotesize
\setlength{\tabcolsep}{2.5pt}
\begin{tabular}{l|c|cc|c}
\Xhline{1.0pt}
 & \multicolumn{1}{c|}{ImageNet} & \multicolumn{2}{c|}{COCO} & \multicolumn{1}{c}{ADE20k} \\
 & top-1   & AP$^\text{box}$ & AP$^\text{mask}$ & mIoU \\
\hline
Swin-T & 81.3 & 50.4 & 43.7 & 44.5 \\
Swin-T w/o Window Shifting  & 80.2 & 47.7 & 41.5 & 43.3 \\
+ Reparam. (window, \#blocks, ratio)  & 81.9 & 50.5 & 43.7 & 45.0 \\
+ GC ViT-T Stem  & 82.2 & 50.7 & 43.9 & 45.2 \\
+ GC ViT-T Down-sampler  & 82.6 & 50.8 & 44.0 & 45.8 \\
\hline
\rowcolor{Gray}
\textbf{+ GC ViT-T Global Self-attention} & \textbf{83.5} & \textbf{51.6} & \textbf{44.6} & \textbf{47.0} \\

\Xhline{1.0pt}
\end{tabular}
  }
    \caption{Ablation study on the effectiveness of various components in GC ViT on classification, detection and segmentation performance.}
    \label{tab:abl_study_cmt}
\end{table}
\textbf{Component-wise Analysis.} As shown in Table~\ref{tab:abl_study_cmt}, we study the role of each component in GC ViT model for classification, detection, instance and semantic segmentation. For simplicity, we start with Swin Transformer as the base model and progressively re-design the components to demonstrate their importance in improving the performance. Firstly, we remove the window shifting and predictably observe significant performance degradation across all tasks. Changing distribution of parameters to our design improves the performance by +1.7, +2.8, +2.2 and +1.7 in terms of accuracy, box AP, mask AP and mIoU. Such reparametrization includes changing the window size, MLP ratio, number of layers to name but a few. Adding the CNN-based stem of GC ViT to the model provides additional improvements of +0.3, +0.2, +0.2 and +0.2 in terms of accuracy, box AP, mask AP and mIoU. In addition, using our proposed downsampler further improves the accuracy, box AP, mask AP and mIoU by +0.4, +0.1, +0.1 and +0.3, respectively. The last two changes demonstrate the importance of convolutional inductive bias and capturing the inter-channel dependencies in our model. Finally, leveraging the proposed global self-attention improves the performance by by +0.9, +0.8, +0.6 and +1.2 in terms of accuracy, box AP, mask AP and mIoU. Hence, this validates the effectiveness of the proposed global self-attention, in particular for downstream tasks with high resolution images such as semantic segmentation in which modeling long-range spatial dependencies is critical.

\subsection{ImageNet-21K}
In Table~\ref{tab:abl-imgnet21K-V1}, we compare the performance of GC ViT-L model which pretrained on ImageNet-21K dataset and finetuned on ImageNet-1K dataset with counterpart approaches. GC ViT-L outperforms Swin-L and CSwin-L by +0.3\% and +0.1\% in terms of Top-1 accuracy respectively, while performing on-par with ConvNeXt-L model. As a result, it validates the effectiveness of the model in large-scale data regimes. 

\begin{table}

\small
\centering
\resizebox{\linewidth}{!}{
\setlength{\tabcolsep}{2.5pt}
  \begin{tabular}{lccc}
    \toprule
    Model   & Param (M)& FLOPs (G)&Top-1 (\%) \\
    \midrule
    Swin-L \citep{liu2021swin} & 197 & 34.5& 86.3 \\
    CSwin-L~\citep{dong2022cswin}  & 173& 31.5& 86.5 \\
    ConvNeXt-L~\citep{liu2022convnet}  & 198& 34.4& \textbf{86.6} \\
    \rowcolor{Gray}
    \textbf{GC ViT-L} &201  &32.6& \textbf{86.6} \\
    \bottomrule
    
  \end{tabular} 
  }
    \caption{Classification benchmarks of \textbf{ImageNet-21K} trained models on \textbf{ImageNet-1K} dataset~\citep{deng2009imagenet}.}
    \label{tab:abl-imgnet21K-V1}

\end{table}

\subsection{Generalizability}
In Table~\ref{tab:abl-imgnetv2}, we have evaluated the performance of GC ViT on ImageNetV2 dataset~\citep{recht2019imagenet} to further measure its robustness. Specifically, we have used different sampling strategies of Matched Frequency and Threshold-0.7. These benchmarks demonstrate the competetive performance of GC ViT on ImageNetV2 dataset and validates its effectiveness in robustness and generalizability.

\subsection{Downsampler Design}
We studied the effectiveness of various downsampler blocks in Table~\ref{tab:abl-sup-downsampler}. The simplest alternative to our design is a pair of convolutional and maxpooling layers. However, it results in a reduction of ImageNet Top-1 accuracy by -0.8. Patch merging is another variant which was introduced in Swin Transformers~\citep{liu2021swin}.

\begin{table}

\small
\centering
\resizebox{\linewidth}{!}{
\setlength{\tabcolsep}{2.5pt}
  \begin{tabular}{lcc}
    \toprule
    Model   & Accuracy-Matched Frequency& Accuracy-Threshold-0.7 \\
    \midrule	 
    GC ViT-XT &71.3  &78.8 \\
    GC ViT-T &73.1  &80.5 \\
    GC ViT-S &73.8  &80.7 \\
    GC ViT-B &74.4  &81.1 \\
    GC ViT-L &74.9  &81.8 \\
    \bottomrule
    
  \end{tabular} 
  }
    \caption{Classiication benchmarks of GC ViT models on ImageNetV2 dataset.}
    \label{tab:abl-imgnetv2}

\end{table}

\begin{table}
\small
\centering
\resizebox{.87\linewidth}{!}{
\setlength{\tabcolsep}{2.5pt}
  \begin{tabular}{lcc}
    \toprule
    Down-sampler   & Architecture& Top-1 \\
    \midrule	 
    Conv  & Conv (s=1), Maxpool&82.7 \\
    Swin &Linear  & 82.9 \\
    \rowcolor{Gray}
    \textbf{GC ViT}&Modified Fused-MBConv (s=2)  & \textbf{83.5}\\
    \bottomrule
    
  \end{tabular} 
  }
    \caption{Ablation study on the effectiveness of down-sampler in GC ViT architecture on ImageNet Top-1 accuracy.}
    \label{tab:abl-sup-downsampler}
\end{table}

However, it reduces the accuracy by -0.6. Finally, our down-sampler which consists of a modified Fused-MBConv block and strided convolution and shows the best result. Importance of the former component is explained by the SE operation which boosts cross channel interaction while keeping number of parameters and FLOPs low. We conclude that our proposed down-sampler is essential to achieve high accuracy as it introduces convolutional inductive bias.

\section{Interpretability}
To provide further insights on interpretability of the proposed global self-attention and query tokens, we demonstrate visualization of the learned attention and Grad-CAM~\citep{selvaraju2017grad} maps  in Fig.~\ref{fig:att_maps}. The associated attention distributions uncovered by the global self-attention modules align with image semantics, and hence act as an informative source for local attention modules. In addition, corresponding Grad-CAM maps demonstrate accurate object localization with most intricate details.

\section{Related work}

\textbf{ConvNet}. Since the advent of deep learning, CNNs~\citep{krizhevsky2012imagenet,SimonyanZ14aVGG,SPPnet,he2016deep,szegedy2016rethinking,huang2017densely,hu2018squeeze} have dominated computer vision benchmarks with SOTA performance. Recently, ConvNeXt~\citep{liu2022convnet} proposed modifications to the architecture of ResNet~\citep{he2016deep}, and achieved competitive benchmarks for classification, detection and segmentation tasks.

\textbf{Transformer}. The ViT~\citep{dosovitskiy2020image} was first proposed as an alternative to CNNs with the advantage of enlarged receptive field, due to its self-attention layers. However, it lacked desirable properties of CNNs such as inductive biases and translation invariance and required large-scale training datasets to achieve competitive performance. Data-efficient Image Transformers (DeiT)~\cite{touvron2021training} introduced a distillation-based training strategy which significantly improved the classification accuracy. 

\textbf{Hybrid}. LeViT~\citep{graham2021levit} proposed a hybrid model with re-designed multi-layer perceptron (MLP) and self-attention modules that are highly-optimized for fast inference. Cross-covariance Image Transformer (XCiT)~\citep{ali2021xcit} introduced a transposed self-attention module for modeling the interactions of feature channels. Convolutional vision Transformer (CvT)~\citep{Wu_2021_ICCV} introduced convolutional token embedding layer and Transformer block in a hierarchical architecture to improve the efficiency and accuracy of ViTs. Conditional Position encoding Vision Transformer (CPVT)~\citep{chu2021conditional} demonstrated improved performance on various tasks such as image classification and object detection by conditioning the position encoding on localized patch token. Tokens-To-Token Vision Transformer (T2T-ViT)~\citep{yuan2021tokens} proposed a transformation layer for aggregating adjacent tokens and establishing image prior by exploiting spatial correlations. Pyramid Vision Transformer (PVT)~\citep{wang2021pyramid} proposed a hierarchical architecture with patch embedding at the beginning of each stage and spatial dimension reduction to improve the computational efficiency. Independently, Swin Transformers~\citep{liu2021swin} also proposed a hierarchical architecture in which self-attention is computed within local windows which are shifted for region interaction. Twins Transformer~\citep{chu2021twins} proposed a spatially separable self-attention with locally-grouped and global sub-sampling modules to improve the efficiency.

\textbf{Global Attention}. Other efforts such as EdgeViT~\citep{pan2022edgevits} in computer vision and BigBird~\citep{zaheer2020big} in NLP have proposed global self-attention in their respective applications. The global attention in GC ViT is fundamentally different than these approaches. For instance, EdgeViT samples representative tokens and only computes sparse self-attention between these representative tokens with reduced feature size. On the contrary, GC ViT computes self-attention between the global queries (not just the token) and local keys and values without any subsampling in their respective local regions. Furthermore, in EdgeViT, only subsampled representative tokens per region interact In the self-attention module; however, in GC ViT, the global queries interact with the entire local regions. Furtermore, BigBird uses a combination of random, window and global attention mechanisms, which is different from the proposed local and global self-attention scheme in GC ViT. BigBird does not have any specific mechanisms for extracting global tokens as the existing tokens or additional special tokens can be specified as global tokens. However, the global tokens in GC ViT are obtained by the query generator via extracting contextual information from the entire input features. Lastly, BigBird employs a set of global tokens which attend to the entire input sequence. However, in GC ViT, the global query tokens attend to local key and value tokens in partitioned windows, since attending to the entire input sequence is not feasible considering the larger size of input features.

\section{Conclusion}
In this work, we introduced a novel hierarchical ViT, referred to as GC ViT, which can efficiently
capture global context by utilizing global query tokens and interact with local regions. We have
extensively validated the effectiveness of our model on various tasks. The proposed GC ViT model
achieves new SOTA benchmarks for image classification across various model sizes on ImageNet-1K
dataset, and surpasses both CNN and ViT-based counterparts by a significant margin. We have
also achieved SOTA or competitive performance for downstream tasks of detection and semantic
segmentation on high-resolution images.

\nocite{langley00}

\bibliography{ref}
\bibliographystyle{icml2023}

\newpage
\appendix
\onecolumn
\section{Appendix}
\renewcommand{\thesection}{\Alph{section}}
\renewcommand\thefigure{S.\arabic{figure}}
\setcounter{figure}{0}
\renewcommand\thetable{S.\arabic{table}}
\setcounter{table}{0}

\subsection{GC ViT Model Configurations}
GC ViT model configurations are presented in Table~\ref{table:arch-spec} describing the choice of internal hyper parameters to obtain models with various compute load and parameter number.

\begin{table*}[h]
\small
\centering
\addtolength{\tabcolsep}{-2pt}
\resizebox{1.\linewidth}{!}{
\begin{tabular}{c|c|c|c|c|c}
 & \begin{tabular}[c]{@{}c@{}}Output Size \\ (Downs. Rate)\end{tabular} & GC ViT-XT  & GC ViT-T &  GC ViT-S &  GC ViT-B \\
\hline
\hline
\multirow{3}{*}{Stem} & \multirow{3}{*}{\begin{tabular}[c]{@{}c@{}}112$\times$112\\ (2$\times$)\end{tabular}} & Conv, C:64, S:2, LN  & Conv, C:64, S:2, LN  &Conv, C:96, S:2, LN   & Conv, C:128, S:2, LN  \\
\cline{3-6}
& & $\begin{bmatrix}\text{F-MBConv}\\\text{C:64}\end{bmatrix}$ $\times$ 1   & $\begin{bmatrix}\text{F-MBConv}\\\text{C:64}\end{bmatrix}$ $\times$ 1    & $\begin{bmatrix}\text{F-MBConv}\\\text{C:96}\end{bmatrix}$ $\times$ 1   & $\begin{bmatrix}\text{F-MBConv}\\\text{C:128}\end{bmatrix}$ $\times$ 1   \\
\hline
\multirow{3}{*}{Stage 1} & \multirow{3}{*}{\begin{tabular}[c]{@{}c@{}}56$\times$56\\ (4$\times$)\end{tabular}} & Conv, C:128, S:2, LN  & Conv, C:128, S:2, LN  & Conv, C:192, S:2, LN   & Conv, C:256, S:2, LN  \\
\cline{3-6}
& & $\begin{bmatrix}\text{LG-SA,}\\\text{C:64, head:2}\end{bmatrix}$ $\times$ 3,  & $\begin{bmatrix}\text{LG-SA,}\\\text{C:64, head:2}\end{bmatrix}$ $\times$ 3,   & $\begin{bmatrix}\text{LG-SA,}\\\text{C:96, head:3}\end{bmatrix}$ $\times$ 3,   & $\begin{bmatrix}\text{LG-SA,}\\\text{C:128, head:4}\end{bmatrix}$ $\times$ 3,   \\
&& F-MBConv, C:128& F-MBConv, C:128&F-MBConv, C:192  & F-MBConv, C:256  \\
\hline
\multirow{3}{*}{Stage 2} & \multirow{3}{*}{\begin{tabular}[c]{@{}c@{}}28$\times$28\\ (8$\times$)\end{tabular}} & Conv, C:256, S:2, LN  & Conv, C:256, S:2, LN  & Conv, C:384, S:2, LN   & Conv, C:512, S:2, LN  \\
\cline{3-6}
& & $\begin{bmatrix}\text{LG-SA,}\\\text{C:64, head:4}\end{bmatrix}$ $\times$ 4,   & $\begin{bmatrix}\text{LG-SA,}\\\text{C:64, head:4}\end{bmatrix}$ $\times$ 4,    & $\begin{bmatrix}\text{LG-SA,}\\\text{C:96, head:6}\end{bmatrix}$ $\times$ 4,   & $\begin{bmatrix}\text{LG-SA,}\\\text{C:128, head:8}\end{bmatrix}$ $\times$ 4,   \\
&& F-MBConv, C:256& F-MBConv, C:256&F-MBConv, C:384  & F-MBConv, C:512  \\
\hline
\multirow{3}{*}{Stage 3} & \multirow{3}{*}{\begin{tabular}[c]{@{}c@{}}14$\times$14\\ (16$\times$)\end{tabular}} & Conv, C:512, S:2, LN  & Conv, C:512, S:2, LN  & Conv, C:768, S:2, LN   & Conv, C:1024, S:2, LN  \\
\cline{3-6}
& & $\begin{bmatrix}\text{LG-SA,}\\\text{C:64, head:8}\end{bmatrix}$ $\times$ 6 ,  & $\begin{bmatrix}\text{LG-SA,}\\\text{C:64, head:8}\end{bmatrix}$ $\times$ 19,    & $\begin{bmatrix}\text{LG-SA,}\\\text{C:96, head:12}\end{bmatrix}$ $\times$ 19,   & $\begin{bmatrix}\text{LG-SA,}\\\text{C:128, head:16}\end{bmatrix}$ $\times$ 19,   \\
&& F-MBConv, C:512& F-MBConv, C:512&F-MBConv, C:768  & F-MBConv, C:1024  \\
\hline
\multirow{3}{*}{Stage 4} & \multirow{3}{*}{\begin{tabular}[c]{@{}c@{}}7$\times$7\\ (32$\times$)\end{tabular}} & Conv, C:1024, S:2, LN  & Conv, C:1024, S:2, LN  & Conv, C:1536, S:2, LN   & Conv, C:2048, S:2, LN  \\
\cline{3-6}
& & $\begin{bmatrix}\text{LG-SA,}\\\text{C:64, head:16}\end{bmatrix}$ $\times$ 5,   & $\begin{bmatrix}\text{LG-SA,}\\\text{C:64, head:16}\end{bmatrix}$ $\times$ 5,    & $\begin{bmatrix}\text{LG-SA,}\\\text{C:96, head:24}\end{bmatrix}$ $\times$ 5,   & $\begin{bmatrix}\text{LG-SA,}\\\text{C:128, head:32}\end{bmatrix}$ $\times$ 5,   \\
&& F-MBConv, C:1024& F-MBConv, C:1024&F-MBConv, C:1536  & F-MBConv, C:2048  \\
\end{tabular}
}
\normalsize
\caption{Architecture configurations for GC ViT. LG-SA and $\text{Conv}$ denotes local, global self-attention and $3\times 3$ convolutional layer, respectively. GC ViT-XT, GC ViT-T, GC ViT-S and GC ViT-B denote XTiny, Tiny, Small and Base variants, respectively.}
\label{table:arch-spec}
\end{table*}

\subsection{Ablation}

\subsubsection{Global Query} We performed ablation studies to validate the effectiveness of the proposed global query. Using the same architecture, instead of global query, we compute: (1) global key and value features and interact them with local query (2) global value features and interact it with local query and key. As shown in Table~\ref{tab:abl_study}, replacing global query may significantly impact the performance for image segmentation and downstream tasks such as object detection, instance segmentation and semantic segmentation. 

\setlength{\tabcolsep}{4pt}
\begin{table}[h]
\centering
\resizebox{.5\linewidth}{!}{
\centering
\footnotesize
\setlength{\tabcolsep}{2.5pt}
\begin{tabular}{l|c|cc|c}
\Xhline{1.0pt}
 & \multicolumn{1}{c|}{ImageNet} & \multicolumn{2}{c|}{COCO} & \multicolumn{1}{c}{ADE20k} \\
 & top-1   & AP$^\text{box}$ & AP$^\text{mask}$ & mIoU \\
\hline
w. Global KV & 82.5 & 49.9 & 41.3 & 44.6 \\
w. Global V  & 82.7 & 50.8 & 42.4 & 45.1 \\
\hline
\rowcolor{Gray}
GC ViT-T & \textbf{83.5} & \textbf{51.6} & \textbf{44.6} & \textbf{47.0} \\

\Xhline{1.0pt}
\end{tabular}
  }
    \caption{Ablation study on the effectiveness of the proposed global query for classification, detection and segmentation.}
    \label{tab:abl_study}
\end{table}

\subsubsection{Effect of Global Context Module}
\begin{figure*}[!t]
\centering
    \includegraphics[width=1.0\textwidth]{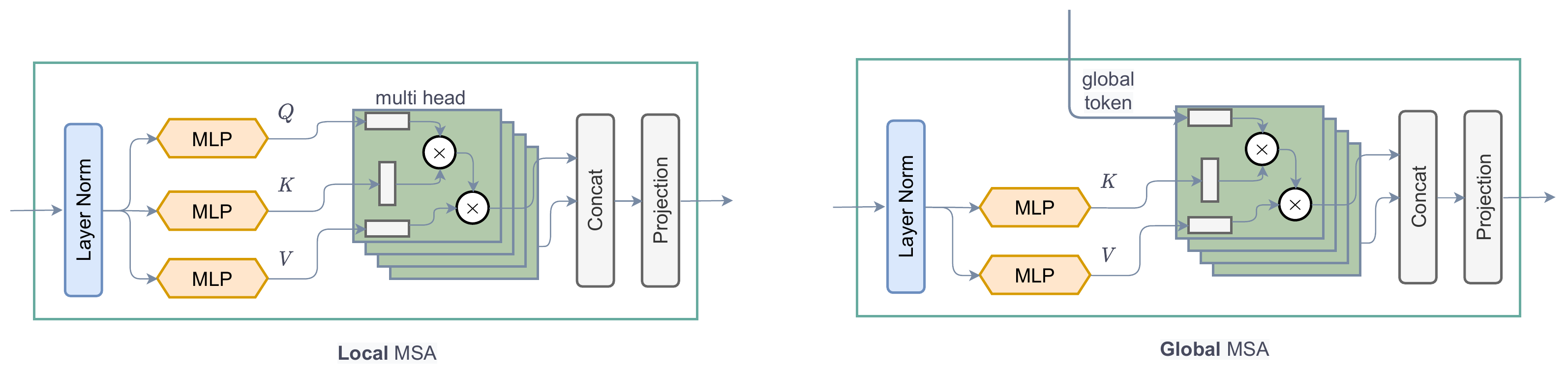}
  \caption{Local and global attention blocks. Global attention block does not compute query vector and reuses global query computed via Global Token Generation. 
  }
  \label{fig:attention_blocks}
\end{figure*}

In Fig.~\ref{fig:attention_blocks}, we illustrate the difference between GC ViT local and global attention blocks. In order to demonstrate the effectiveness of Global Context (GC) self-attention module, we use Swin Transformers as the base model and add our propoped GC module. In this analysis, we remove the window shifting operation from Swin Transformers, since GC module is capable of modeling cross-region interactions. As shown in Table~\ref{tab:abl-sup-gcmovule}, addition of GC module improves the ImageNet Top-1 accuracy by $+0.9\%$ and $+0.7\%$ for Swin Transformers Tiny and Small variants respectively.

\begin{table}
\small
\centering
\resizebox{.30\linewidth}{!}{
\setlength{\tabcolsep}{2.5pt}
  \begin{tabular}{lcc}
    \toprule
    Model  & Added Component& Top-1 \\
    \midrule	 
    Swin-T  & None &81.3 \\
    Swin-T &GC Module  & 82.2 \\
    Swin-S &None & 83.0 \\
    Swin-S &GC Module  & 83.7 \\
    \bottomrule
    
  \end{tabular} 
  }
    \caption{Ablation study on the effectiveness of Global Context (GC) module in Swin Transformers architecture on ImageNet Top-1 accuracy.}
    \label{tab:abl-sup-gcmovule}

\end{table}

\subsubsection{EMA and Batch Size} We also used used Exponential Moving Averages (EMA) and observed slight improvement in terms of ImageNet TOp-1 accuracy. Furthermore, the performance of the model across different batch sizes were stable as we did not observe significant changes. Table~\ref{tab:ema} demonstrates the effect of EMA and batch size on the accuracy of a GCViT-T model.

\vspace{-2mm}

\begin{figure*}[!t]
\centering

\resizebox{0.85\linewidth}{!}{
\begingroup
\renewcommand*{\arraystretch}{0.3}
\begin{tabular}{c}

  \includegraphics[width=1\linewidth]{figures/imgs.jpg} \\ 
  \small{(a) Original images from ImageNet-1K validation set.} \\[3pt]

\includegraphics[width=1\linewidth]{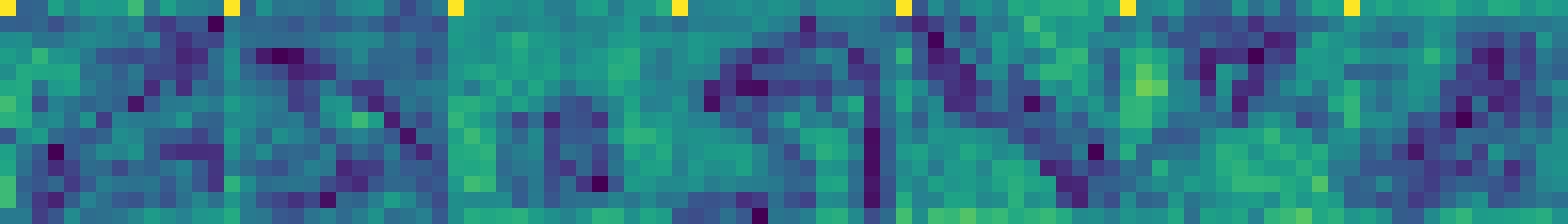} \\
 \small{(b) Learned \textbf{global query} tokens.}
\end{tabular}
\endgroup
}
\caption{Visualization of : (a) input images (b) learned global query token feature maps.}
\vspace{-2mm}
\label{fig:att_maps_supp}
\end{figure*}

\setlength{\tabcolsep}{4pt}
\begin{table}[h]
\centering
\resizebox{.5\linewidth}{!}{
\setlength{\tabcolsep}{2.5pt}
  \begin{tabular}{lcccc}
    \toprule
    Model&Local Batch Size   & Global Batch Size&EMA&  Top-1 \\
    \midrule	 
    GC ViT-T&32  & 1024&No&83.45 \\
    GC ViT-T&128  & 4096&No&83.46 \\
    GC ViT-T&32  & 1024&Yes&83.47\\
    \hline
    \rowcolor{Gray}
   GC ViT-T&128  & 4096&Yes&83.48\\
    \bottomrule
    
  \end{tabular} 
  }
    \caption{Ablation study on the effect of EMA and batch size on GC ViT-T ImageNet Top-1 accuracy.}
    \label{tab:ema}
\end{table}

\subsection{Training Details}
For image classification, GC ViT models were trained using four computational nodes with 32 NVIDIA A100 GPUs. The total training batch size is $1024$ ($32$ per GPU) for GC ViT-S, GC ViT-B, GC ViT-L and $4096$ ($128$ per GPU) for GC ViT-XXT, GC ViT-XT and GC ViT-T. On average, each model required $32$ hours of training with the specified hyper-parameters as indicated in the paper. All classification models were trained using the \verb|timm| package~\citep{rw2019timm}. Object detection and instance segmentation models as well as semantic segmentation models were trained using one computational node with 8 NVIDIA A40 GPUs using a total batch size of $16$, hence a batch size of $2$ per GPU. Detection and instance segmentation models were trained using \verb|mmdetection| \citep{chen2019mmdetection} package and on average required $56$ hours of training.  Semantic segmentation models were trained using \verb|mmsegmentation|~\citep{mmseg2020} package, and on average required $34$ hours of training.


\subsection{Interpretability}
In Fig.~\ref{fig:att_maps_supp}, we illustrate the learned global query token maps and demonstrate their effectiveness in capturing long-range contextual representations from different image regions.


\end{document}